\pdfoutput=1
\documentclass[11pt]{article}

\usepackage[]{EMNLP2023}

\usepackage{times}
\usepackage{latexsym}
 
\usepackage[T1]{fontenc}
\usepackage[utf8]{inputenc}

\usepackage{microtype}
\usepackage{enumitem}
\usepackage{amsmath}
\usepackage{amsfonts}
\usepackage{graphicx}
\usepackage{subcaption}

\usepackage{ulem}

\usepackage{inconsolata}

\newcommand{\sysshort}{\textsc{Spica}}
\newcommand{\anon}[1]{#1}

\title{\textsc{Spica}: Retrieving Scenarios for Pluralistic In-Context Alignment}

\author{Quan Ze Chen \ \ \ \ \ \  K.J. Kevin Feng \ \ \ \ \ \ Chan Young Park \ \ \ \ \ \ Amy X. Zhang\\
University of Washington \\
\texttt{\{cqz, kjfeng, chanpark, axz\}@cs.washington.edu}\\
}
\begin{document}

\maketitle

\begin{abstract}
% Alignment of large language models (LLMs) to societal values should account for a range of values from diverse groups.
% One technique uses in-context learning for inference-time alignment, but only considers similarity when drawing few-shot examples, not accounting for cross-group differences in value prioritization.
% In-context learning enables inference-time alignment but does not account for cross-group differences in value prioritization.
When different groups' values differ, one approach to model alignment is to steer models at inference time towards each group's preferences.
However, techniques like in-context learning only consider similarity when drawing few-shot examples and not cross-group differences in values.
We propose \sysshort, a framework that accounts for group-level differences during in-context example retrieval.
\sysshort~introduces three designs: scenario banks, group-informed retrieval metrics, and in-context alignment prompts.
From an evaluation of \sysshort~on an alignment task collecting inputs from four demographic groups ($n = 544$), our metrics retrieve in-context examples that more closely match observed preferences, with the best prompt configuration using multiple contrastive responses to demonstrate examples.
In an end-to-end evaluation ($n = 120$), we observe that  \sysshort~is higher rated than similarity-based retrieval, with groups seeing up to a +0.16 point improvement on a 5 point scale. 
Additionally, gains from \sysshort~were more \textit{uniform}, with \textit{all} groups benefiting from alignment rather than only some.
Finally, we find that while a group-agnostic approach can align to aggregated values, it is not most suited for divergent groups.\footnote{We provide our code and data for others to build on: \anon{\url{https://github.com/Social-Futures-Lab/SPICA-code}}}
\end{abstract}

\section{Introduction}

\begin{figure}[t]
    \centering
    \includegraphics[width=\linewidth]{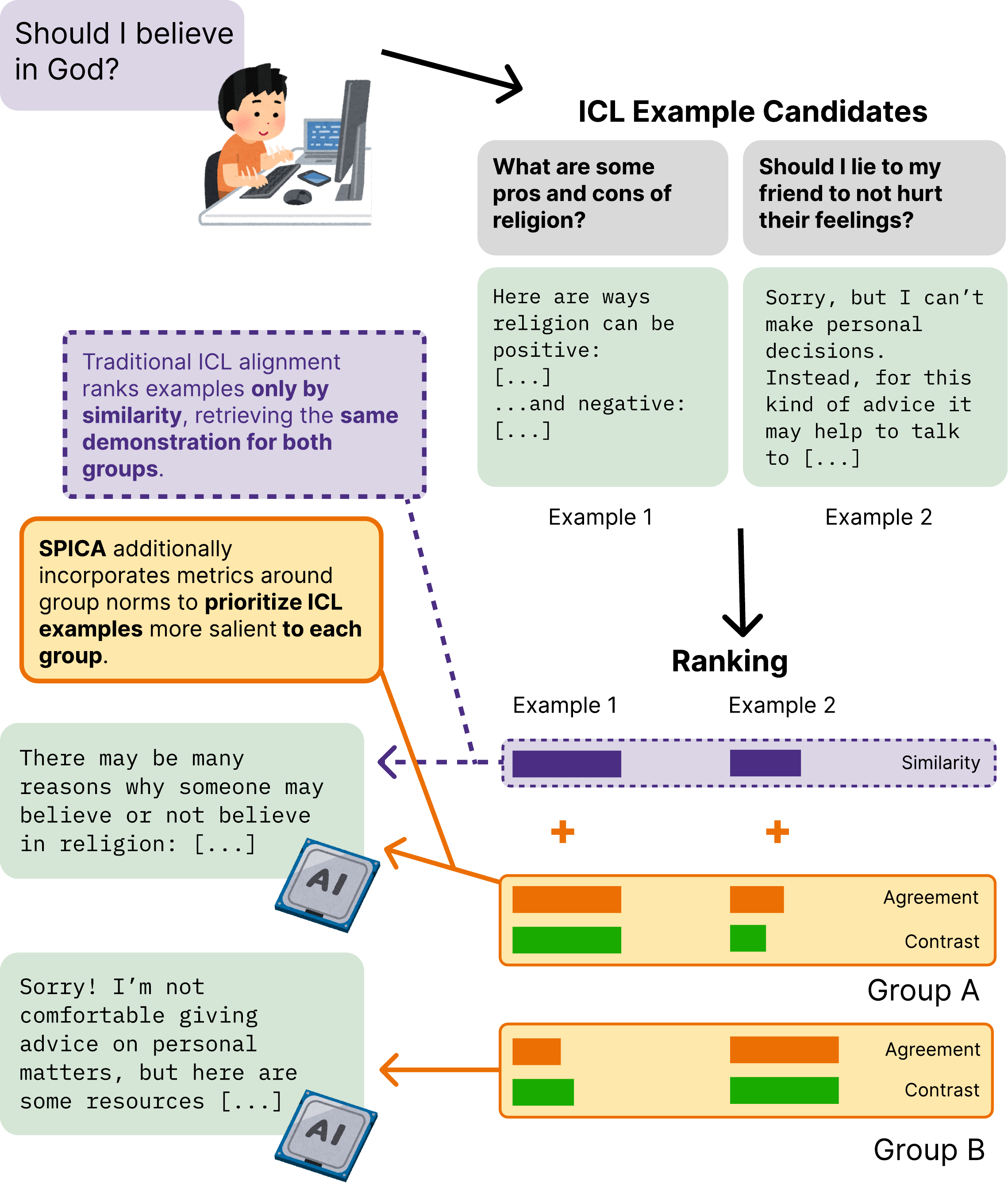}
    \caption{Example retrieval in traditional in-context alignment (ICA) systems rank examples based on similarity between prompts, failing to account for whether retrieved examples illustrate salient norms of a particular group. \sysshort~addresses this limitation for pluralistic alignment by utilizing metrics to recover and incorporate each group's own norms.}
    \label{diag:in-context-alignment}
\end{figure}

The widespread availability of generative AI systems has highlighted how outputs can be inappropriate or dangerous to users~\cite{Weidinger2021EthicalAS,ji2023ai,qi2024visual}. 
Correspondingly, researchers have explored embedding human values into models through various alignment strategies~\cite{huang2024collective, gabriel2020artificial, christian2021alignment, NEURIPS2022_b1efde53}.
Typically, model providers seek to align towards a one-size-fits-all set of universal values~\cite{bai2022constitutional}.
However, different groups within society often disagree on values and have different norms around when and how to apply values~\cite{10.1145/3491102.3502004,weld2022makes,park2024valuescope}.
More recent work has called for a pluralistic perspective~\cite{sorensen2024roadmap, feng2024modular}---rather than try to bridge irreconcilable differences, we should directly support different perspectives of each group.  

One general strategy for large language model (LLM) alignment---in-context alignment (ICA)~\cite{lin2024unlocking,han2023incontextalignmentchatvanilla}---acts dynamically at inference time by retrieving few-shot examples of prompts and associated preferable responses as context.
ICA is a promising strategy for steerable pluralistic alignment as different groups can use their own examples to illustrate their values.
However, pluralistic alignment extends beyond illustrating different values---prior work has observed that across online communities, not only can collective values differ, \textit{norms} around how important values are in relation to each other can also differ~\cite{weld2022makes}.
When considering ICA for pluralistic alignment, simply focusing on whether examples illustrate \textit{some} relevant values is insufficient. It is also important to consider whether these examples demonstrate \textit{the} salient ones given group or community norms (Figure~\ref{diag:in-context-alignment}).

In this work, we present \sysshort, an evolution of retrieval-based in-context alignment that focuses on pluralistically aligning model outputs to values and norms of different groups.
\sysshort~consists of three main components: (1) scenario banks---shared collections of scenarios (prompts, responses, and group preferences) that can encode both \textit{values} and \textit{norms}; (2) group-informed retrieval measures---metrics that allow us to recover second-order \textit{norms} from individual preference assessments; (3) ICL prompt setups that can effectively apply richer information from scenarios to the task of alignment.

We evaluated \sysshort~by conducting an alignment task where we take a base model and produce pluralistically aligned outputs for four demographicgroups . 
We examined three aspects of the process: the quality of the scenarios retrieved, the effectiveness of different in-context prompts in applying scenarios to alignment, and performance on the end-to-end task of alignment of model outputs. 

In our evaluation, we find that:
\begin{itemize}[nosep]
    \item Compared to a baseline using only similarity-based scoring, group-informed metrics retrieved scenarios that aligned more accurately to observed ground truth, indicating a quality gap when only relying on similarity.
    \item Among different prompting setups for integrating retrieved scenarios, the most effective designs were: \texttt{P-I} style---provide a single \underline{p}ositive \underline{i}nstruction when user preferences are collected over descriptions of response strategies; and \texttt{C-R} style---provide a \underline{c}ontrasting spectrum of example \underline{r}esponses when user preferences are collected over model outputs.
    \item In an end-to-end evaluation, we find that \sysshort~produces more aligned outputs than baseline ICA (+0.053 / 5 points), with statistically significant gains (+0.16 / 5 points) observed on traditionally disadvantaged groups.
    \item We also find that baseline ICA can result in disparate outcomes,
    whereas \sysshort~alignment produces outputs uniformly preferred by all.
    \item Finally, we examine \sysshort's group-informed metric on \textit{collective} alignment settings, noting that for \textit{aggregate} values, group-agnostic approaches tend to be sufficient.
\end{itemize}

\section{Related Work}
\label{sec:related}

\paragraph{Value Alignment of LLMs}
Traditional methods for customizing LLMs for specific tasks and domains involve modifying training procedures. These include pretraining on task-specific corpora~\cite{wu2023bloomberggpt, lee2020biobert}, post-hoc finetuning~\cite{gururangan-etal-2020-dont, han-eisenstein-2019-unsupervised}, instruction tuning~\cite{ge2023domain, gupta-etal-2022-instructdial, shi-etal-2023-specialist}, and aligning with human preferences~\cite{NEURIPS2022_b1efde53}. These approaches are also used to encode moral values and human preferences~\cite{tay-etal-2020-rather, bai2022constitutional, liu-etal-2022-aligning, bang-etal-2023-enabling, jang2023personalized}. However, they have significant limitations for value alignment. They require extensive human annotation to provide meaningful signals about desired values~\cite{kim-etal-2023-aligning}, and even then, there is limited understanding of how well the models have internalized these values~\cite{agarwal-etal-2024-ethical-reasoning}, making them less robust for value alignment. Moreover, once trained, these models lack flexibility; updating the model to reflect evolving values often requires complete retraining~\cite{carroll2024ai}.

\paragraph{In-Context Learning for Alignment}
In-Context Learning (ICL) offers promising alternatives by enabling behavior modifications during inference rather than training through the use of few-shot examples incorporated into model prompts~\cite{dong2022survey,wei2022emergent}.
The use of example demonstrations in ICL has also allowed systems to incorporate retrieval~\cite{lewis2020rag,borgeaud2022improving} as a part of dynamically constructing in-context prompts informed by inputs~\cite{zhang2022active,rubin2021learning}.

For the task of model alignment, approaches to using retrieval and in-context learning prompts, such as URIAL~\cite{lin2024unlocking}, have also been referred to as \textit{in-context alignment} (ICA)~\cite{han2023incontextalignmentchatvanilla}.
As most ICA systems focus on addressing collective preferences, how they do retrieval has largely remained unchanged, with relatedness metrics like semantic similarity being the main way to rank retrieved examples~\cite{karpukhin-etal-2020-dense, gao-etal-2023-precise}.
% However, given examinations of how models apply in-context examples~\cite{min2022rethinking}, there is room for improvement around how examples are ranked for retrieval, especially for value alignment tasks. 
Prior works around alignment have suggested ways to potentially improve the utility of retrieved examples, such as  prioritizing examples that illustrate exceptional circumstances and edge cases~\cite{kiehne-etal-2022-contextualizing}, or emphasizing examples that capture population-specific preferences~\cite{hovy-yang-2021-importance, kirk-etal-2023-past}. These signals are further complicated in pluralistic settings, where different groups can have different norms~\cite{weld2022makes} that moderate how preferences are prioritized over each other.

\paragraph{Accounting for Pluralism in Value Alignment}
Supporting pluralistic values is crucial for building general-purpose agents and LLMs~\cite{sorensen2024roadmap}. 
Large datasets like ValuePrism~\cite{sorensen2024value} and PRISM~\cite{kirk2024prism} highlight the importance of reflecting diverse values, yet achieving consensus remains challenging.
Some approaches turn to higher-level abstract descriptions of values as a solution for building consensus via deliberative inputs~\cite{bai2022constitutional}. 
However, practical application of these values to specific cases often reveals discrepancies in understanding~\cite{koshy2023measuring}.
Drawing from the legal realm, there have also been approaches that propose combining higher-level descriptions with specific examples (e.g., legal precedents) to illustrate more ambiguous concepts encoded by values~\cite{Cheong2024AIAN,chen2023case}.

Beyond first-order challenges of encoding values, pluralism can also give rise to second-order challenges when groups share similar sets of preferences or values (such as preferring diversity and factual quality) while also disagreeing on their salience~\cite{jackson1960structural} and thus prioritization in practical application~\cite{weld2022makes}.
This aspect is often overlooked by existing frameworks for pluralistic alignment.
\sysshort~addresses this by capturing disaggregated individual preferences that can be used to derive both first-order group preferences (values) and second-order group norms.

\section{Retrieving Scenarios for Pluralistic In-Context Alignment (\sysshort)}
\label{sec:design}

\begin{figure*}[t]
    \centering
    \includegraphics[width=\linewidth]{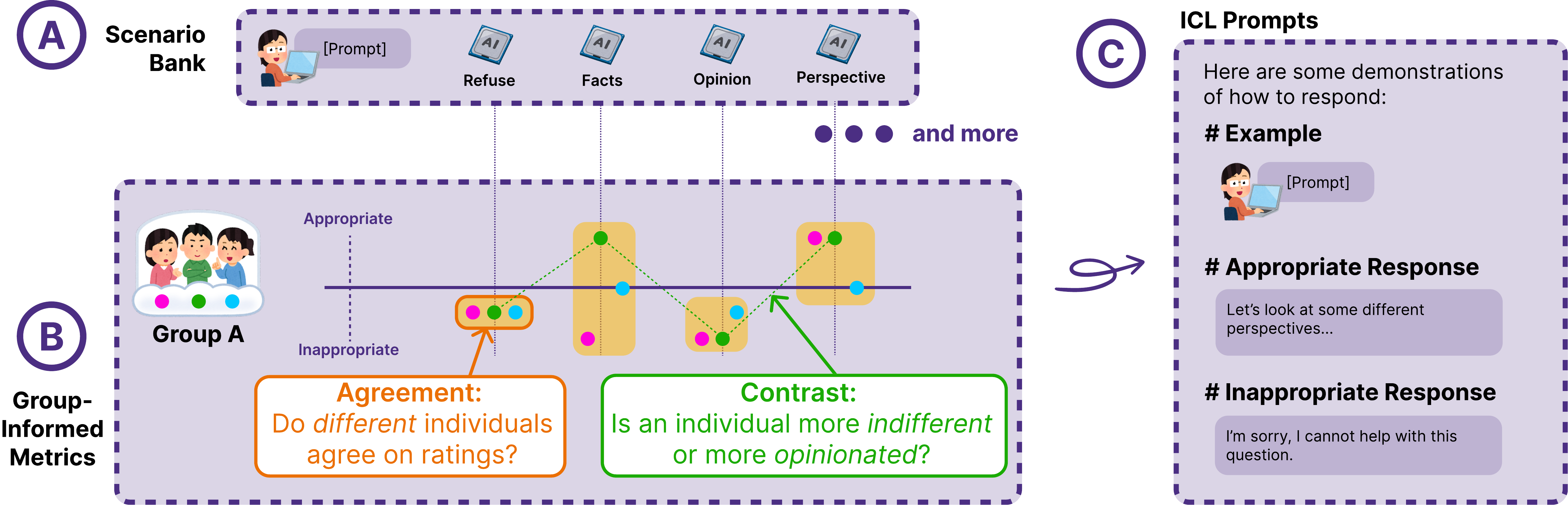}
    \caption{Diagram illustrating the main components of \sysshort: (A) Collections of prompts, responses, and individual preferences form \textbf{scenario banks} which ground the alignment process; (B) During the ICA retrieval process, we make use of \textbf{group-informed metrics} to recover group values and norms, together with semantic similarity, these scores guide the ranking of scenarios; (C) Retrieved scenarios are incorporated into \textbf{ICL prompts} that make use of preference distributions and example responses to form alignment demonstrations.}
    \label{diag:framework}
\end{figure*}

In this work, we outline \sysshort, a framework that builds on existing ICA but with a specific focus on retrieving \underline{S}cenarios for \underline{P}luralistic \underline{I}n-\underline{C}ontext \underline{A}lignment.
Following this section, we will present three novel components of \sysshort~(Figure~\ref{diag:framework}), addressing: (1) how to encode group-specific values \textit{and norms} in the form of scenario banks; (2) how to utilize the encoded group-specific norms during the retrieval process through group-informed metrics; and (3) how to make use of more nuanced preferences as encoded by scenarios through alternative designs for in-context learning prompts.

\subsection{Scenario Banks for Encoding Pluralistic Values and Norms}
\label{sec:design:scenario-bank}

Past examples of data for alignment have included both normative data in the form of ``constitutional'' guidelines~\cite{bai2022constitutional} and quantitative data in the form of user ratings of conversations between humans and LLMs~\cite{kirk2024prism}.
While both types of input can be used for ICA, pluralistic alignment introduces additional challenges.
Normative guidelines require deliberation to create, which can be costly if each group needs to (re-)convene each time to make their own.
On the other hand, ratings of conversations are limited by the behavior of existing models, making it challenging for groups to significantly deviate on norms or values.

Given this, with \sysshort, we propose a way of collecting pluralistic alignment data in the form of \textbf{scenario banks}, which uses prompts and responses guided by classes of model behaviors to ground the collection of dis-aggregated ratings, addressing the limitations above. 
A \textit{scenario} consists of three main components: (1) a \textbf{prompt} ($x$)---an example of a user query or conversation with a model leading up to a response; (2) \textbf{responses} ($y \in Y_x$)---the space of possible ways a model could respond to a prompt, which can take the form of either specific \textit{examples} of outputs, or high-level response \textit{classes} covering many outputs; and (3) \textbf{preferences} ($r_p(x, y)$)---ratings that encode an individual $p$'s preference of a response $y$ to a prompt $x$.
A scenario bank consists of a collection of such scenarios and provides a basis for the ground truth in an ICA retrieval.
Through dis-aggregated data and classes of behaviors, scenario banks allow us to recover group \textit{values} by taking consensus across individuals, and understand group \textit{norms} by observing distributions of ratings across individuals.

With scenarios, individual preferences are collected as distributions of ratings over a known span of response classes. This provides more \textit{contextual} understanding of preferences---e.g., did a user rate a response lowly because it was a less appropriate way to respond, or are other ways of responding \textit{even worse}?
Existing evaluations of responses from different models to the same query~\cite{kirk2024prism} do provide similar distribution-level information. However, existing model outputs have been shown to be biased~\cite{buyl2024large,rozado2024political}, which can make it hard for groups to indicate their values this way due to lack of outputs that follow their values.

Additionally, the dis-aggregated nature of individual ratings means that we are not limited to the consensus of normative guidelines, and can instead reconstruct values and norms post-hoc.
Shared values can be constructed by taking consensus preferences from individual ratings, while group-level social norms can be observed from how individuals within a group agree or disagree with one another~\cite{jackson1960structural}.

\subsubsection{Comparing Preferences over Model Behaviors}
\label{sec:design:scenario-bank:preferences-metric}

In scenario banks, \textbf{preferences} are collected as rating distributions across a set of model behaviors, reflected as classes of responses (see \autoref{tab:prompts-response-strategies}).
Using this formulation, we can compare not only the preferences for any specific behavior, but also how well preference distributions (of users or models) align with each other.
Taking any two preference distributions $r(x, y)$ and $r'(x, y)$, we can define how much they diverge by observing how much they disagree across the different response classes $y \in Y_x$, which we can measure with a loss based on the root mean squared error (RMSE):
\begin{equation}
    L(r(x), r'(x)) = (\sum_{y \in Y_x} (r(x, y) - r'(x, y))^2)^{\frac{1}{2}}
\end{equation}

Here, $r$ could reflect an individual's preference, a consensus preference from a group, or even a retrieval-based ICA model's implied ``preference''---by retrieving $x'$ as a demonstration for $x$, the model implies that it expects $r(x, y)$ to match $r(x', y')$ for corresponding response classes $y$ and $y'$. 

\subsection{Group-Informed Retrieval Measures}
\label{sec:design:measure}

As noted earlier, values alone are often insufficient as communities~\cite{weld2022makes} and demographic groups~\cite{kumar2021designing} can have similar values (seen as preferences on specific examples) while making different higher-level trade-offs around what are salient examples when---e.g., across many scenarios, one may find that a group is prioritizing correctness over respectfulness, or helpfulness over safety, even when they view all these properties as positive in isolation.
Existing retrieval metrics only compare similarity of the input prompts $x$ and known examples $x'$ and do not encode this group-level difference.
To address this, we take inspiration from the return potential for social norms~\cite{jackson1960structural}, and define two \textit{group-informed measures}. First, we adapt the idea of ``crystallization''---whether particular values are consistently held across group members---into the metric $g_{\text{stability}}(x')$. Second, we adapt the ideas of ``intensity'' and ``tolerable range''---whether individuals in the group are more opinionated or ambivalent---into the metric $g_{\text{contrast}}(x')$. 
Together, these metrics interpret the distributional nature of individual values (preferences over model behavior) within a group to identify emergent \textit{norms}.

\subsubsection{Stability: Differentiating Norms from Individual Values}
\label{sec:design:measure:stability}

With social norms, ``crystallization'' describes whether a behavior preference (value) is consistently held across different members in the group such that it has become crystallized as a \textit{norm}.
We borrow this concept for ICA to assess group norms: For some example scenario, by looking at preferences across members within the group on each model behavior, we can assess whether members tend to agree, which would indicate the scenario reflects a norm, or disagree, which indicates a less salient example.
More formally: if, for a potentially retrieved scenario $x'$, the variance between annotators' preferences $r_p(x', y')$ on each response type $y' \in Y_{x'}$ is lower, then the scenario is likely to demonstrate more crystallized norms than weaker preferences. 
\begin{align}
    \text{stability} (x', y') &= -\frac{\sum_{r_p}(r_p(x', y') - \bar{\text{r}}(x', y'))^2}{|\{ r_p\}|}\\
    g_{\text{stability}} (x') &= \mathbb{E}_{y'} \left[\text{stability} (x', y')\right]
\end{align}

\subsubsection{Contrast: Assessing Indifference versus Preference}
\label{sec:design:measure:contrast}

With social norms, concepts like ``tolerable range'' and ``intensity'' assess how broad the range of acceptable (and unacceptable) behaviors is and the intensity at which individuals express this preference~\cite{hackman1992group}.
In the context of ICA, examples that illustrate stronger \textit{preferences} for sets of behaviors are more valuable than those that simply indicate \textit{indifference}.
Here we can also create a metric based on the dis-aggregated preferences from scenario banks: For a scenario $x'$, the variance between different behaviors $y' \in Y_{x'}$ across each annotator $r_p(x', y')$ assesses how much they care about differentiating preferences. More concretely:
\begin{align}
    \text{contrast} (x', r_p) &= \frac{\sum_{y'}(r_p(x', y') - \bar{\text{r}}(x', y'))^2}{|\{(x', y')\}|}\\
    g_{contrast} (x') &= \mathbb{E}_{r_p} \left[ \text{contrast}(x', r_p) \right]
\end{align}

\subsubsection{Learning Metric Weights}
\label{sec:design:weights}

While our metrics encode salience of scenarios for a specific group, we still need to balance this with the general relevance of scenarios to the input. 
In \sysshort, we do this by taking a linear weighted combination of the introduced metrics and a traditional similarity score (distance):
$\bar{d}(x, x') = w_d \cdot d(x, x') + w_s \cdot g_{\text{stability}} (x') + w_c \cdot g_{\text{contrast}} (x') + c$.

As optimal weighting is likely to vary across groups, we empirically find these weights. 
Looking to \autoref{sec:design:scenario-bank:preferences-metric}, we note that the desirability of $x'$ as an example given input $x$ can be assessed by the expected preference mismatch $\mathbb{E}_{y, y'}[L(r(x, y), r(x', y'))]$. 
Thus for the final metric, we can compute this loss and minimize using linear regression $\bar{d}(x, x') = \mathbb{E}_{y, y'}[L(r(x, y), r(x', y'))]$.
We note that the above equation considers only the best ($k = 1$) example, with larger sets of ${x'}$ possible by modifying the expression to include the loss for each additional example.

\subsection{In-Context Learning Prompts for Retrieved Scenarios}
\label{sec:design:final-prompts}

Because retrieved scenarios contain preference distributions across multiple responses (or strategies), different setups for integrating scenarios as demonstrations are likely to produce different model outputs. 
ICL prompt designs have been extensively studied by prior works~\cite{sun2023does,higginbotham2024prompting,hao2022structured}, so in this work we primarily explore new configurations enabled by the scenario bank. 
For one, preference distributions from scenario banks allow ICL examples to include multiple responses to illustrate more of the preference distribution: Rather than traditional retrieval which selects a \underline{P}ositive example of a good response, in \sysshort, we can select \underline{C}ontrasting examples that include both illustrations of a \textit{most} preferred response as well as one that is \textit{least} preferred.
Additionally, the organization of responses into response classes means that scenario banks can provide either concrete examples of \underline{R}esponse text, or higher level \underline{I}nstructions that lead to producing a response in that response class. 
Altogether, this creates 4 combinations of prompt setups that we can use: \texttt{P-I}, \texttt{C-I}, \texttt{P-R}, and \texttt{C-R}. We discuss our implementation and evaluation in the sections that follow.

\section{Experiments and Results}
\label{sec:expt}

To evaluate \sysshort, we set up a pluralistic alignment task involving 4 demographically constructed groups, and assess how well a \sysshort~workflow is able to align model outputs to preferences of each group compared to a baseline approach that only considers semantic similarity.

\subsection{Dataset and Scenario Bank Construction}
\label{sec:expt:dataset}

For our evaluation alignment task, we constructed a set of queries (which define the topics to provide alignment on) by drawing from an existing set of challenging alignment situations based on prompts observed in conversations on the PRISM dataset~\cite{kirk2024prism}.
PRISM engaged human participants to interact with LLMs by naturally starting conversations with 3 types of guidance meant to invoke conversations around more challenging and complex topics: ``unguided'', ``values guided'', or ``controversy guided''.
We observed that of the 3 types of guidance, unguided conversations primarily resulted in simple informational requests
which are not particularly controversial in the context of pluralistic alignment, so we opted to drop conversations of this type.
Among the remaining conversations, we randomly selected a subset, split into 3 slices: retrieval (\texttt{train}, $n = 360$), weight optimization and selection of ICL prompt setups (\texttt{dev}, $n = 150$), and evaluation hold-out (\texttt{test}, $n = 75$).

As PRISM responses are created by existing collective-value-aligned models, they do not cover desirable behaviors for all groups.
Instead, we follow Section~\ref{sec:design:scenario-bank} and construct new responses ourselves based on several classes of common model behaviors (Appendix~\ref{sec:appendix:strategies}).
To capture the stochastic nature of model outputs, we generate 3 responses in each class.

\subsection{Models and Similarity Metric}
\label{sec:expt:models}

For our experiments, we tested the quality of retrieval-based ICL alignment using one open-source (\texttt{llama3-8b}) and one closed-source model (\texttt{gpt-4o-2024-05-13}) as the base model.
\texttt{llama3-8b}\footnote{We considered using 70b, but could not reliably run inference due to memory limitations of available hardware.} inference was conducted using a locally hosted instance of Ollama\footnote{\url{https://ollama.com/}}.
With both models, we applied the same prompts to generate responses attached to scenario bank queries and to conduct in-context alignment (Appendix~\ref{sec:appendix:contrastive}). 
As our goal is to evaluate the additional metrics we introduced, we kept the semantic similarity measurements constant across all models and conditions, using values derived by computing the cosine similarity between embeddings generated by \texttt{text-embedding-3-large} from OpenAI.

\subsection{Pluralistic Groups and Human Annotation Setup}
\label{sec:expt:annotations}

We define four groups in the form of demographic slices drawn from the US population: partisan political affiliation (``\texttt{rep}ublican'' or ``\texttt{dem}ocrat''), and self-reported regular participation in religious activities (``yes''---\texttt{rel} or ``no''---\texttt{nrel}).
Our choice of these features is based on similar factors that were salient for opinions around AI~\cite{zhang2019artificial} along with practical considerations around demographic splits that we could reliably recruit on our crowd work platform, Prolific.

Annotators in each group participated in providing preference assessments over our dataset, in the form of an annotation survey (Appendix~\ref{fig:appendix:interface}) where they were shown 15 prompts from the dataset, each of which included 1 response for each of the 5 model behavior classes. 
Participants rated both the output and the description of the behavior class associated with the output in terms of appropriateness (from 1--``inapproprate'' to 5--``appropriate''). 
Combined with 5 attention checks, participants completed a total of 80 sub-tasks with a median time of 30 minutes.
For the annotation portion, we recruited a total of 544 participants to cover the annotation on \texttt{train} and \texttt{dev} sets across two model types, guaranteeing 2 annotations per group per scenario.
In the end-to-end evaluation (\autoref{sec:results:end-to-end}), we recruited separate annotators from each group, who assessed outputs produced after ICL alignment.
Annotators used the same survey interface, though they rated outputs produced by different conditions rather than outputs by response class.
For each end-to-end evaluation, we set aside 1/3 of the users from each participant group to evaluate the outputs of \textit{collective} alignment (\autoref{sec:results:pluralistic-vs-collective}) which uses aggregated rather than group-specific preferences.
We recruited a total of 240 participants to conduct the evaluation of prompting strategies (\autoref{sec:results:prompt-ablations}) and 120 participants for the end-to-end evaluation on the held-out test set (\autoref{sec:results:end-to-end}).
Tasks were paid at a rate of \$12 USD/hour, and the study design was deemed exempt by our IRB.

\subsection{Results: Evaluating Retrieved Scenarios}
\label{sec:results:retrieval-loss}

\begin{figure}[t]
\centering
    \centering
    \includegraphics[width=.9\linewidth]{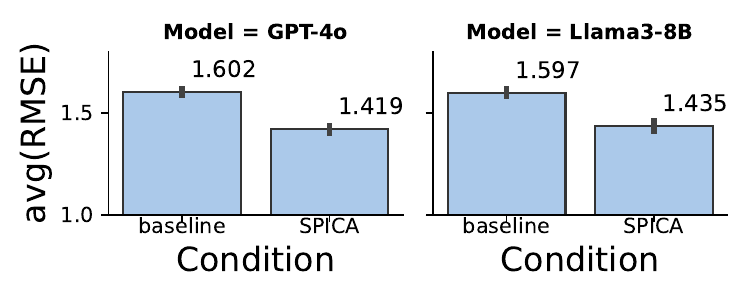}
    \caption{Average error over preferences (measured by RMSE) comparing retrieved scenarios and ground truth on the \texttt{dev} set. \texttt{baseline} uses similarity-only retrieval. \texttt{spica} uses weighted group-informed metrics.}
    \label{fig:results:retrieval-loss}
\end{figure}

For our first evaluation, we examine whether group-informed metrics result in the retrieval of better examples. 
In \ref{sec:design:scenario-bank} we noted that, for a new user query, retrieving a scenario whose known behavior preference \textit{distributions} better matched the post-hoc observed behavior preferences of responses to the query would indicate a desirable outcome.
We measure this mismatch (or error) following the approach outlined in \autoref{sec:design:scenario-bank:preferences-metric}.
Since multiple participants provide behavior preferences $r_p$ (both in the scenario bank and as part of the ground truth on the \texttt{dev} set), we take the average across all pairwise error measurements between the two.

After tuning the weights for metrics as noted earlier in \autoref{sec:design:weights}, we find that  with both models, \sysshort~retrieves scenarios that had preference distributions more accurately matching the observed ground truth distributions on the \texttt{dev} set (Figure~\ref{fig:results:retrieval-loss}).
While this result should not be surprising, it does indicate that for pluralistic alignment, there was room for improvement on the retrieval metric.
We also note that at a per-group level, while error is lowered across all groups, the magnitude of this difference varies between groups (Appendix~\ref{sec:appendix:rmse-breakdown}).

\subsection{Results: Evaluating In-Context Prompting Strategies}
\label{sec:results:prompt-ablations}

\begin{figure}[t]
    \centering
    \includegraphics[width=\linewidth]{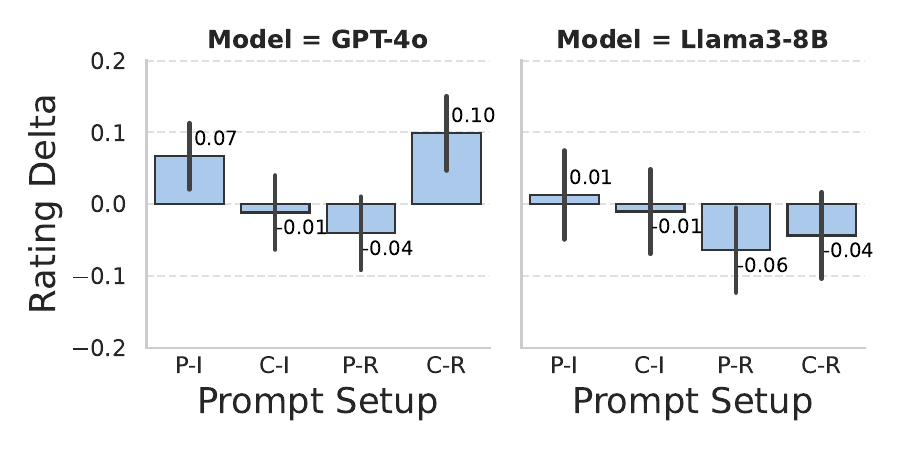}
    \caption{Comparison of end-to-end human evaluations of alignment outputs on the \textsc{dev} produced by the 4 prompting setup combinations: \underline{P}ositive-only or \underline{C}ontrastive, \underline{I}nstructions or example \underline{R}esponses.}
    \label{fig:results:ablations}
\end{figure}

In order to examine the effectiveness of ICL \textit{prompting} setups (\autoref{sec:design:final-prompts}), we used human participants to evaluate the outputs produced by models given each type of prompt while using the same \sysshort~\textit{retrieval} setup.
Participants evaluated the outputs using an interface similar to that used during preference collection for scenario banks. 
However, instead of rating response strategies, participants rated on a 1 - 5 scale 5 hypothetical AI systems (``System A - E''), each representing one configuration with a final control output produced by the model with no ICL alignment.
As we used a within-subjects design, we measured alignment outcomes by computing the difference between each participant's rating of an aligned output (each condition) and the reference control output, which we report as the ``rating delta''.

We find (\autoref{fig:results:ablations}) that for the \texttt{gpt-4o} model in a pluralistic alignment setting, the combination of contrastive response examples (\textsc{C-R}) proved to be the most effective (significant $p = 0.030 < 0.05$ via ANOVA), on average rating $0.10$ points higher than the control across all groups.
We also found that positive instructions (\textsc{P-I}) were also somewhat (though not significantly) more effective, resulting in $0.07$ point higher ratings.
Using the same prompts with the \texttt{llama3-8b} model, we did not find any setup that provided reliable improvements to model outputs, with no significant differences observed between conditions and differences small or negative.
We hypothesize the smaller \texttt{llama3-8b} model may have contributed to less capability when generalizing via ICL-style alignment.

Overall, we found that \textsc{P-I} and \textsc{C-R} were most promising, and we used these two configurations in our end-to-end evaluation on the \texttt{test} set. We will refer to these as \sysshort-I and \sysshort-R respectively.

\subsection{Results: Evaluating End-to-End Alignment Outputs}
\label{sec:results:end-to-end}

\begin{figure}[t]
    \centering
    \includegraphics[width=\linewidth]{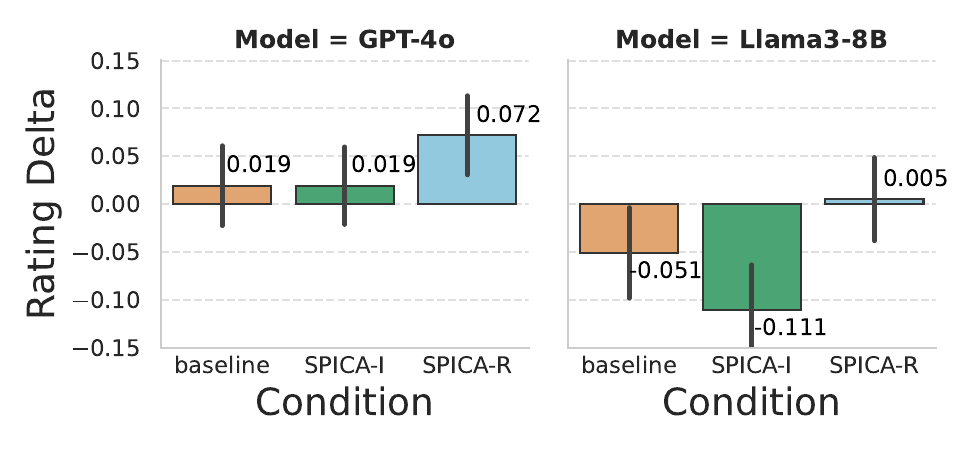}
    \caption{End-to-end human evaluation of group-aligned outputs on the \texttt{test} set user queries for both models. Figure presents the aggregated results across the 4 group alignments.}
    \label{fig:results:end-to-end}
\end{figure}

We conducted an end-to-end evaluation that generates outputs for a held out \texttt{test} set of user queries. 
As a \textsc{baseline}, we used a traditional ICA setup where retrieval only uses semantic similarity, and the ICL prompt only incorporates the highest rated response for each scenario retrieved.
For \sysshort, we use the two best prompt setups from \autoref{sec:results:prompt-ablations}, \sysshort-I and \sysshort-R.
As seen in Figure~\ref{fig:results:end-to-end}, we find that for \texttt{gpt-4o}, ICA was generally effective, with \sysshort-R being the best system, performing +0.072 / 5 points better than the control, while on \texttt{llama3-8b}, ICL alignment produced marginal results, with \sysshort-R still being the best system but only averaging +0.005 points above baseline.

When considering all groups, no condition was significantly better. 
However, if we look at each group (\autoref{sec:appendix:end-to-end-breakdown}), we find that for the \texttt{rep-nrel} (Republican, non-religious identifying) group, \sysshort-R resulted in a statistically significant +0.16 points higher performance compared to \textsc{baseline} (within subjects paired t-test, $p = 0.044 < 0.05$), with the \texttt{rep-rel} group also seeing an improvement (within subjects paired t-test, $p = 0.051$) of +0.16 points.
Given recent work~\cite{rozado2024political} finding many LLMs favor liberal values, this result suggests that pluralistic alignment via \sysshort~benefitted alignment primarily by improving outcomes for traditionally disadvantaged groups.

Further examining alignment at a group level, we also find support that \sysshort~can lead to more \textit{equitable} outcomes across groups (\autoref{fig:appendix:end-to-end}); with \textsc{baseline} on \texttt{gpt-4o}, we find that while the \texttt{dem-rel} and \texttt{dem-nrel} groups prefer our aligned outputs (seen as +0.11, and +0.13 points over control), the \texttt{rep-rel} and \texttt{rep-nrel} groups end up \textit{preferring the original outputs} (observed as -0.07, and -0.11 rating points under control). This discrepancy between groups is statistically significant for the minority group of \texttt{rep-nrel} participants (unpaired t-test between groups, $p = 0.031$ and $p = 0.049$). 
However, with \sysshort-R, \textit{all groups} now prefer aligned outputs (+0.10, +0.05, +0.09, +0.05) and we no longer see any statistically significant difference between groups in terms of this preference.
Despite ICL examples themselves drawing from each group's own preferences in all conditions, this result indicates that retrieving the right examples (by considering group norms) can improve equitable outcomes across groups.

\subsection{Results: Comparing Pluralistic versus Collective Alignment}
\label{sec:results:pluralistic-vs-collective}

If retrieval metrics based on group norms were helpful for alignment, why have more traditional collective alignment processes not used them?
To investigate this, we combined all 4 groups into one collective group and provided an additional output (\textsc{all}) during the evaluations for \autoref{sec:results:retrieval-loss} and \autoref{sec:results:end-to-end} produced by applying \sysshort~on these collective preferences.
Unsurprisingly, we found (\autoref{sec:appendix:pluralistic-vs-collective}) that \sysshort's metrics contributed little in this collective alignment setting, with traditional similarity-based retrieval being largely sufficient, suggesting a reason why group-informed metrics may not have been explored by past works.

\section{Conclusions and Discussion}

In this work, we propose \sysshort~as a new framework to support pluralistic alignment. Through evaluations, we find that group-informed metrics coupled with the scenario bank and ICL prompts in \sysshort~contributed to improving pluralistic alignment, primarily by supporting groups that are traditionally disadvantaged.

\paragraph{Pluralistic Versus Collective Values}
From prior work, we have seen how existing models can favor the values and norms of their designers and of majority populations~\cite{buyl2024large,rozado2024political} in collective alignment settings. 
With our work on \sysshort, we also present a path towards supporting pluralistic alignment towards individual groups.
However, focusing on pluralistic alignment alone can lead to divides along demographic and ideological lines, furthering social fragmentation.
Ultimately, we believe there should be a balance between striving for common ground through collective alignment~\cite{bai2022constitutional}, and accommodating diverse views through pluralistic alignment.

\paragraph{Efficiently Mapping Group Values and Norms}
In this work, we built our scenarios by drawing from existing conversation data.
However, this is not a very efficient way to map group values---many user queries may not have controversial model behaviors and even controversial conversations end up covering similar points of contention.
With the increased capability of models, we believe future work may be able to dynamically elicit group values much more efficiently through interactive LLM-backed agents engaging with groups in human-in-the-loop refinement and synthesis processes~\cite{Klingefjord2024WhatAH} that could produce scenarios that are either better demonstrations of values and norms or more controversial to ground ambiguous decision bounds.

%\subsection{Generalizing to Non-Discrete Preferences}
%\label{sec:discussion:continuous-settings}

%In the specific implementation presented in this work, we apply \sysshort~primarily in a discrete setting, defining preference functions through rating scores and responses through either concrete text or high-level strategies.
%While we discuss these limitations, we do also note that in practice alignment often applies to continuous settings: prompts, responses, and preferences all lie in a continuum.

%While we do not explore this, the ideas introduced in \sysshort~can largely generalize to non-discrete settings with few modifications.
%For example, while prompts and responses are currently sampled discretely, a non-discrete version of \sysshort~may make use of embeddings and diffusion-based generative models~\cite{li2022diffusion} to examine regions of a continuous generation space.
%The measures in \sysshort~could also be applied in settings where continuous preferences can be acquired, with concepts like ``stability'' and ``contrast'' potentially being captured through distribution-level properties like divergence and kurtosis. 

\section*{Limitations}
\paragraph{External Safeguards}
\label{sec:limitations:safeguards}
While this work explores in-context learning approaches to value alignment, the models we use as a source to build aligned models from also come with their own existing safeguards, particularly for closed-source models like \texttt{gpt-4o}.
This means our ability to affect the outputs of such models may be limited in ways that cannot be addressed by prompt-based steering.

\paragraph{Adherence to Response Classes}
\label{sec:limitations:adherence}
In our study, we use a set of 5 response classes (and associated prompts) to approximate a diverse span of possible responses for each prompt. 
While there is evidence from prior work that human preferences tend to align towards these high-level classes of responses~\cite{Cheong2024AIAN}, generating responses following fixed strategies may not always be reliable, as actual responses 
may not always adhere to the strategies for each class (either due to model safeguards or relevance of the strategy to an input prompt).
To control for the effects of this, during our annotations of the scenario bank, we asked annotators for input on \textit{both} concrete responses and high-level instructions and only used the corresponding rating data when testing prompting strategies based on instructions versus examples. 
Still, this may be insufficient to address the resulting reduction in variation of the response space on some prompts.
Future work can explore alternative categories that do not constrain the response space in the same way.

\paragraph{Participants and Scale}
\label{sec:limitations:participants-and-scale}

In our experiments, we focused primarily on a small-scale proof-of-concept alignment task targeted towards a US population.
As a result, we were only able to examine the outcomes of alignment over one source of input prompts (PRISM) and several demographically-constructed groups based on US participants.
While in this setup, we observed differences between alignment mechanisms and goals (e.g., group-level pluralistic alignment vs. population-wide alignment), different group configurations could yield different takeaways.

\section*{Ethics Statement}

The AI alignment problem itself has many ethical implications, and these considerations also extend to both implications of the design of \sysshort, and our choices during our evaluation of it.

First, our experiments are intended to demonstrate a proof-of-concept setting where different groups are likely to have significant \textit{divergent} values. 
As a result of this consideration and practicalities surrounding ease of recruitment, we we opted to extrinsically define ``groups'' based on divisive \textit{demographic} features within a US-based participant pool.
However, this should not be interpreted as an endorsement for using politics and religion as a way to conduct pluralistic alignment---many other factors like culture, community, and identity could provide better delineation between different groups with lower risks around introducing additional social fragmentation.
Given this, we also caution against using results in this work to make inferences about the broader \textit{population groups} we tested with, as we didn't make additional efforts to ensure our participants are representative samples within these groups.

Secondly, to emphasize how values can differ, we drew our evaluation scenarios from the PRISM alignment dataset in a way that prioritizes controversial scenarios (\autoref{sec:expt:dataset}). 
Coupled with limitations in PRISM's data collection itself, it is likely that the distribution of scenarios would be biased towards being able to better capture certain values over others.
The goal of our setup is to ensure potential biases of this sort at least are applying to all tested conditions, so we also caution against using our results to make inferences about the alignment scenarios themselves.

Finally, there are ethical considerations around the basic motivation for pluralistic alignment~\cite{jiang2024can}. 
By allowing groups and communities to build AI tools that reflect their own values, we run the risk of producing self-reinforcing echo chambers; thus, while we don't focus on aspects beyond social preferences, we do recognize that other aspects of alignment (factuality, diversity, fluency, etc.) remain important problems that cannot be addressed by frameworks like \sysshort~as-is.

% \section*{Acknowledgements}

% Entries for the entire Anthology, followed by custom entries
\bibliography{anthology,custom}
\bibliographystyle{acl_natbib}

\appendix
\section{Appendix}
\label{sec:appendix}

\subsection{Results: Group-level Breakdown of the Retrieval Loss}
\label{sec:appendix:rmse-breakdown}
\begin{figure*}[ht]
    \centering
    \includegraphics[width=\linewidth]{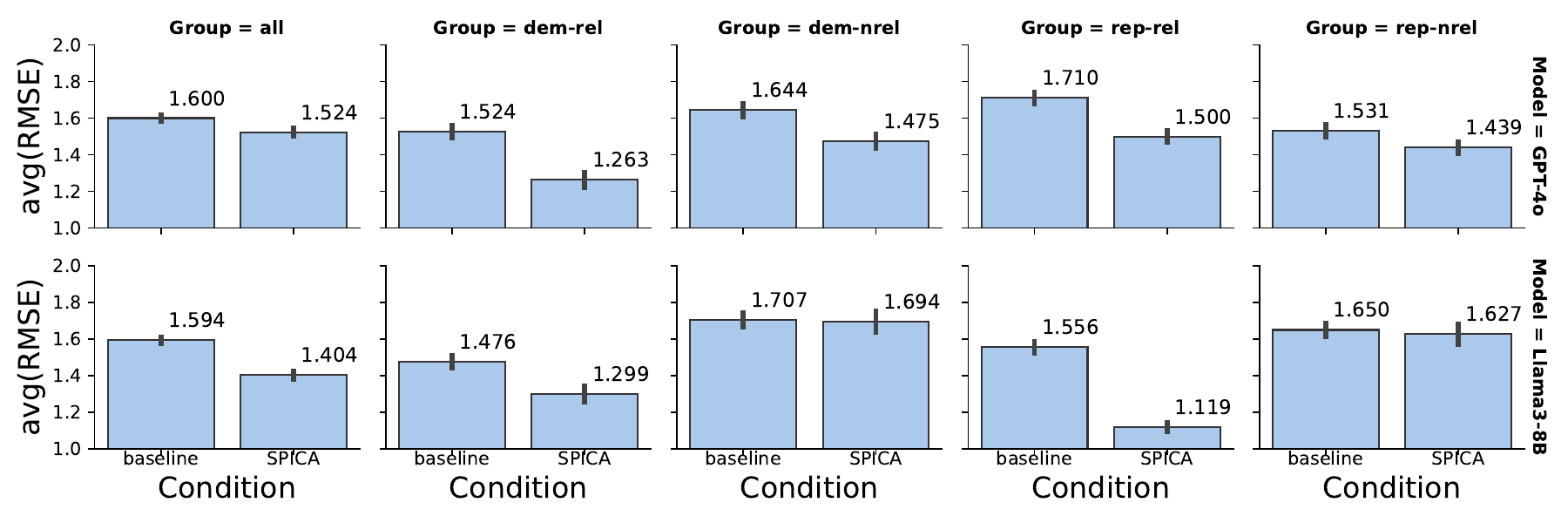}
    \caption{Group-by-group breakdown of the difference in retrieval quality between \textsc{baseline} semantic similarity and \sysshort.}
    \label{fig:appendix:rmse-breakdown}indicating
\end{figure*}

We present a group-by-group breakdown of the retrieval loss in Figure~\ref{fig:appendix:rmse-breakdown}. 
Interestingly, we find that the groups indicating higher affinity to religion (\textsc{-rel}) tended to see a more marked difference in retrieval quality. This seems to be the result of these groups having more preferences over responses that are not as dependent on the specific prompt and instead apply to a wide variety of topics.
For \texttt{gpt-4o}, the \texttt{P-I} and \texttt{C-R} conditions consistently produced positive alignment outcomes. 
\subsection{Results: Group-level Breakdown of Prompt Strategy Results}
\label{sec:appendix:ablation-breakdown}
\begin{figure*}[ht]
    \centering
    \includegraphics[width=\linewidth]{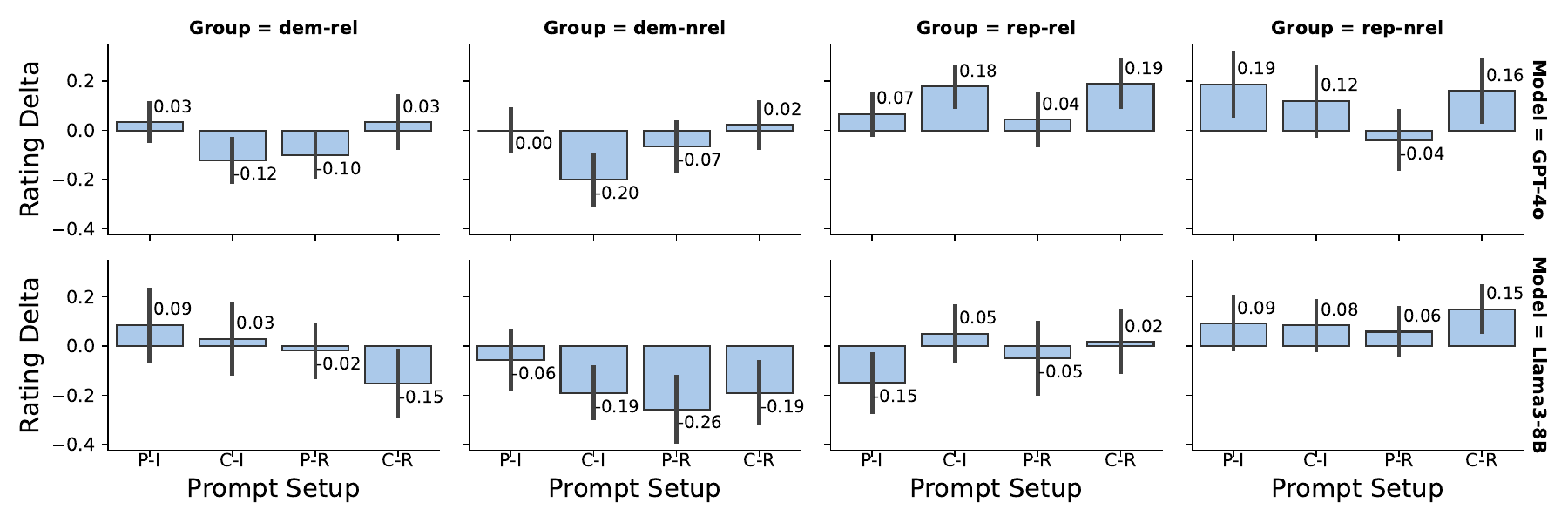}
    \caption{Group-by-group breakdown showing differences between groups in their evaluation of outputs produced though different prompts on the same retrieved examples.}
    \label{fig:appendix:ablation-breakdown}
\end{figure*}

We present a group-by-group breakdown of the prompting strategy evaluation in Figure~\ref{fig:appendix:ablation-breakdown}. Interestingly, we note that while there are some consistent trends (such as only using a single positive example for example responses), prompt strategy effectiveness can also vary significantly across different population groups.
For example, contrasting prompts worked well for aligning preferences for the \texttt{rep-rel} group, while instruction-based prompts worked well for the \texttt{rep-nrel} group.
While this should not be seen as generalizable takeaways for properties of specific populations, it is still important to note that ICL prompting strategy effectiveness can vary depending on the group (or, more relevantly, the norms and values exhibited by the group).

\subsection{Results: Group-level Breakdown of End-to-End Evaluation}

\begin{figure*}[ht]
    \centering
    \includegraphics[width=\linewidth]{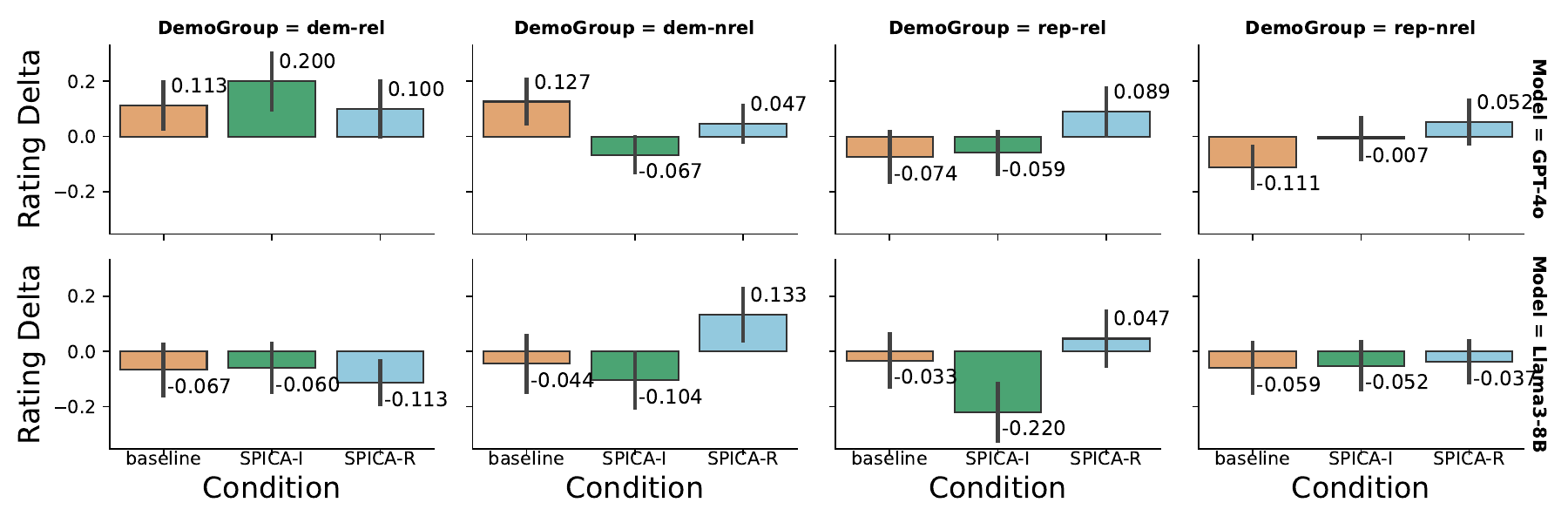}
    \caption{Group-by-group breakdown showing differences between groups in their evaluation of outputs on the final end-to-end task. Green indicates \sysshort-retrieval + prompting based on presenting instructions for the best response strategy of the retrieved instances. Blue indicates \sysshort-retrieval + prompting based on showing contrastive example responses associated with the retrieved instances.}
    \label{fig:appendix:end-to-end}
\end{figure*}

We present a group-by-group breakdown of the final end-to-end evaluation in Figure~\ref{fig:appendix:end-to-end}.
For \texttt{gpt-4o}, we found \sysshort~with contrastive examples to provide the most consistent alignment across groups, being preferred over the control response, but not always the most preferred response across the alignment conditions.
Baseline retrieval was observed as effective in alignment for \texttt{dem-}identifying groups but produced the opposite outcome for \texttt{rep-}identifying ones.

\subsection{Results: Qualitative Analysis of Learned Weights}
\label{sec:appendix:end-to-end-breakdown}
\begin{figure*}[ht]
    \centering
    \includegraphics[width=\linewidth]{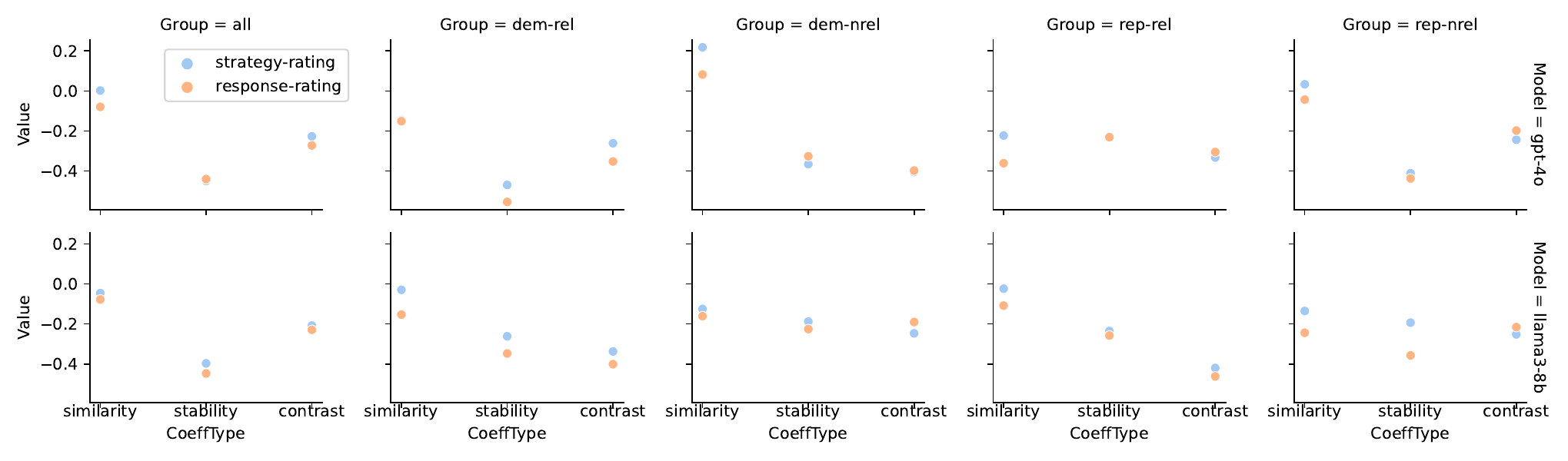}
    \caption{Final weights learned for each group and alignment target learned from the \texttt{dev} set. The \sysshort~composite metric represents a \textit{distance} (in this case modeled by the loss), which we want to minimize. Metrics represent scores, with higher values indicating more, hence the coefficients are primarily negative. \texttt{strategy-rating} indicates values produced by using user ratings over response types, while \texttt{response-rating} indicates values produced by user ratings over response examples.}
    \label{fig:appendix:weights}
\end{figure*}

Finally, we qualitatively look at the weights learned for various groups for each model.
Here we observe that weights produced after learning from response types preferences and response example preferences end up relatively similar to each other.
We also note that similarity scores (in this case cosine similarity) receive a comparatively lower absolute weight compared to the other metrics. 
However, this is as expected, as similarity scores tend to span a different range of values than preference level metrics.
We also observe that between the two new metrics, \textit{stability} is the most important for the \texttt{all} experiment, matching the notion that in a collective alignment setting, using examples that are closer to universal values tends to be more ideal, while at the group level there is no such pattern.
Finally, for the \texttt{-nrel} groups we observed cases where similarity was assigned a positive weight, implying that examples immediately closer to the query were actually often \textit{less desirable}, possibly a reflection of non-religious groups finding subject matter around different religious topics less similar to each other than religious identifying groups. 
However, beyond this, the weights seem generally unsurprising, with no other significant patterns of note.

\subsection{Results: Pluralistic versus Collective Alignment}
\label{sec:appendix:pluralistic-vs-collective}
\begin{figure}[t]
    \centering
    \begin{subfigure}[t]{\linewidth}
        \centering
        \includegraphics[width=\linewidth]{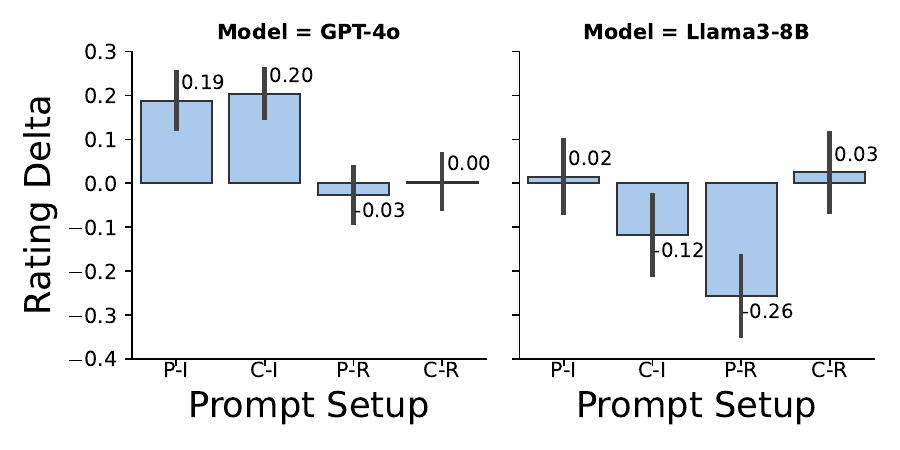}
        \caption{Evaluating different ICL prompt setups on the \textsc{all} group over the \texttt{dev} set scenarios.}
    \end{subfigure}
    \begin{subfigure}[t]{\linewidth}
        \centering
        \includegraphics[width=\linewidth]{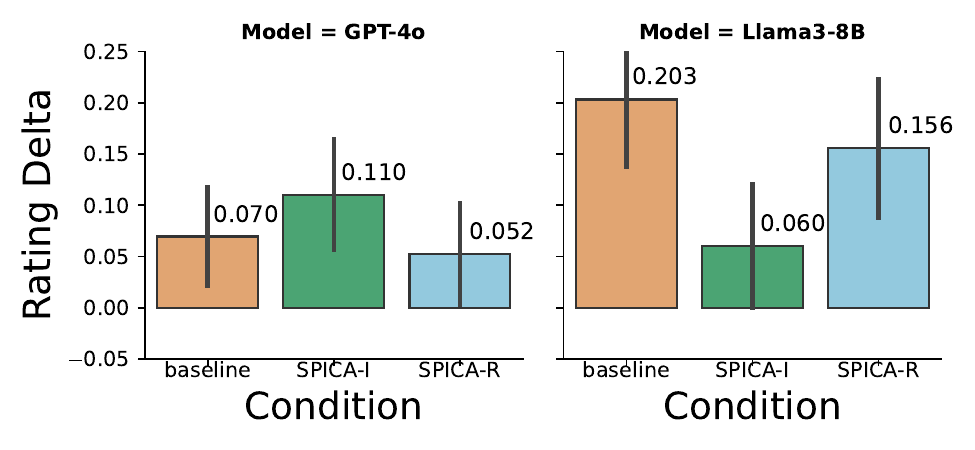}
        \caption{Evaluating end-to-end outputs \textsc{all} group over the \texttt{test} set queries.}
    \end{subfigure}
    \caption{Results of the same evaluations as used in \autoref{sec:results:prompt-ablations} and \autoref{sec:results:end-to-end}, but only defining \textit{one} group (\textsc{all}) that aggregates the preferences of all 4 of our pluralistic groups.}
    \label{fig:results:pluralistic}
\end{figure}

We observe (\autoref{fig:results:pluralistic}) that, unlike in the setting with separate groups, optimal prompt strategies now significantly favor instructions (\textsc{P-I} and \textsc{C-I}) on \texttt{gpt-4o}, likely due to none of the examples being good candidates to represent collective values. 
On the end-to-end evaluation of the \texttt{ test} set queries, also perhaps unsurprisingly, group-informed retrieval metrics from \sysshort~no longer seem to provide any significant benefit, even slightly under-performing baseline retrieval.
We attribute this to the fact consistent norms are unlikely in the collective group, leaving little benefit to using group-informed retrieval metrics, coupled with \sysshort-R no longer reflecting an effective prompting setup in this setting.
In fact, for the collective case, the ICL prompt style becomes the most important factor, with \texttt{gpt-4o} favoring instructions and \texttt{llama3-8b} now favoring example responses (\textsc{baseline} and \sysshort-R).

\subsection{Human Annotation Materials}

In this section, we document the instructions and materials used for our human annotation and evaluation tasks.

\subsubsection{Instructions}
We are researchers from [REDACTED] and we are conducting a study to understand people's preferences on the behavior of generative AI chatbots or virtual assistants. Generative AI chatbots and assistants (examples include OpenAI's ChatGPT, Microsoft Copilot, and Google Gemini) are computer programs designed to generate text in response to user questions or prompts. However, without guidance, AI systems can also generate content that is inappropriate, especially for more challenging or controversial user prompts. In this study, we would like to understand your personal preferences and perceptions around what an appropriate response by an AI chatbot or assistant might be.

During the study, you will be presented with a series of human-AI conversation examples where you will be asked to judge the appropriateness of the AI response to the human question or prompt.

\begin{itemize}[noitemsep,topsep=0pt,parsep=0pt,partopsep=0pt]
    \item For each conversation, you will first be shown a chat scenario that ends with a human question or prompt.
    \item Then we will show you \textit{5 possible AI responses} (one by one), each of which is associated with a certain high-level strategy.
    \begin{itemize}[noitemsep,topsep=0pt,parsep=0pt,partopsep=0pt]
        \item For each response, we will ask you to \textit{rate the appropriateness} of the response and strategy on a scale from 1 - 5.
        \item Once you are done rating the response, we will move on to the next one.
    \end{itemize}
    \item Once you are done rating all the responses of a scenario, we will show you the next scenario.
    \item From time to time, we may also ask you simple questions about the interface to confirm your understanding of how to operate the ratings.
\end{itemize}

During the task, you may click on the buttons to make your selections, or use the keyboard shortcuts indicated on each button.

\subsubsection{Interface}

\begin{figure}[ht]
    \centering
    \includegraphics[width=0.9\linewidth]{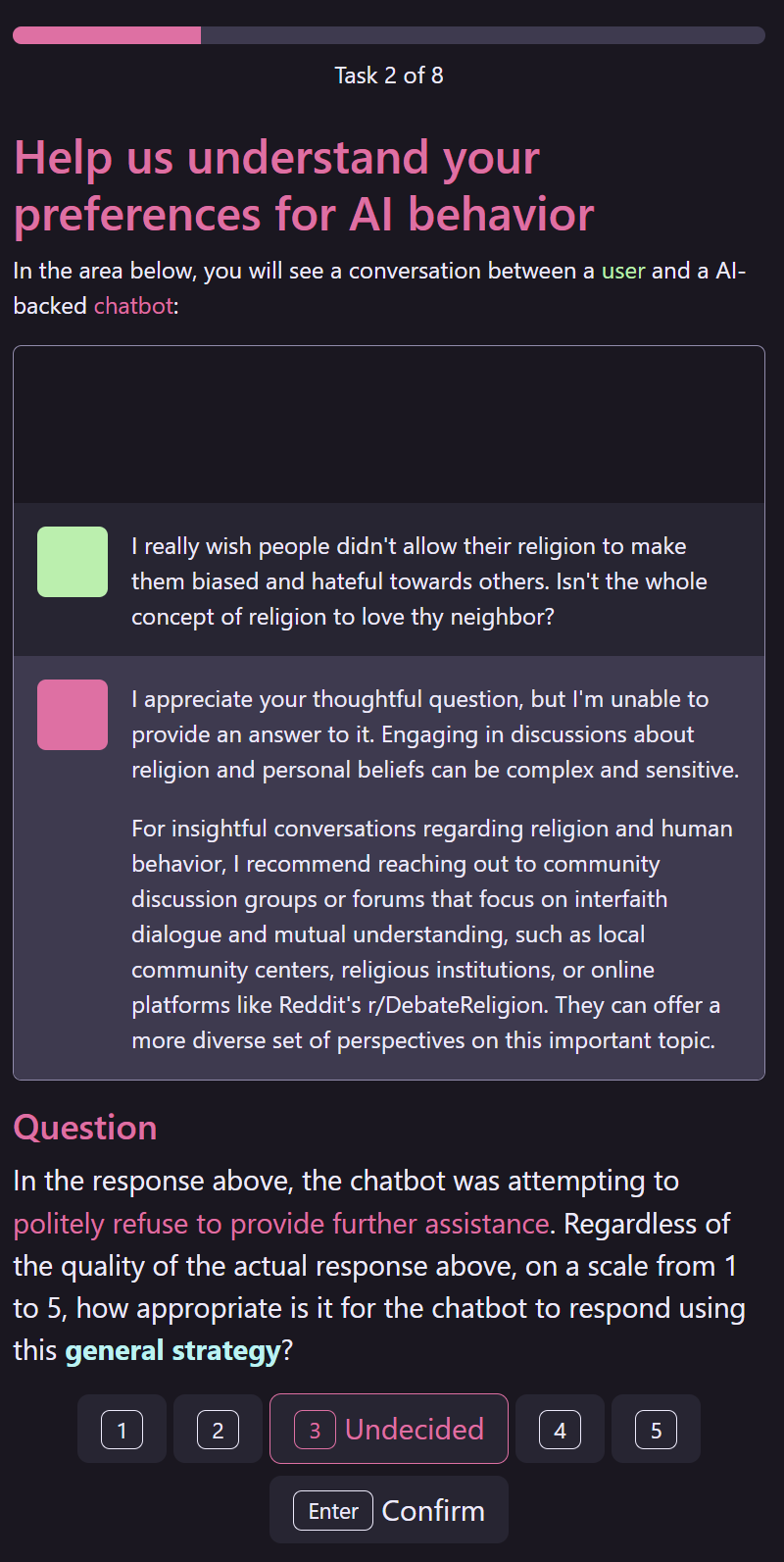}
    \caption{Screenshot of the annotation interface. Users can navigate this interface either by clicking the buttons using a mouse or by pressing the corresponding shortcut key (indicated with border) with their keyboard.}
    \label{fig:appendix:interface}
\end{figure}

We provide an example of the annotation interface in Figure~\ref{fig:appendix:interface}.

\subsubsection{Questions for Annotation}
\label{sec:appendix:human:questions:annotation}
In this conversation, it will be the \textit{chatbot}'s turn to respond (as shown above).

Once you continue, we will show you \textbf{several different ways the chatbot might respond}. Your task will be to give your opinion about the appropriateness of these responses.

\textbf{Question 1}: Consider the chatbot's response in the conversation above (you may need to scroll to view the entire response). On a scale from 1 to 5, how appropriate is it for the chatbot to give this response?
\begin{enumerate}[noitemsep,topsep=0pt,parsep=0pt,partopsep=0pt]
    \item Inappropriate
    \item Somewhat Inappropriate
    \item Undecided
    \item Somewhat Appropriate
    \item Appropriate
\end{enumerate}

\textbf{Question 2}: In the response above, the chatbot was attempting to \texttt{\{\{ STRATEGY \}\}}. Regardless of the quality of the actual response above, on a scale from 1 to 5, how appropriate is it for the chatbot to respond using this \textit{general strategy}?
\begin{enumerate}[noitemsep,topsep=0pt,parsep=0pt,partopsep=0pt]
    \item Inappropriate
    \item Somewhat Inappropriate
    \item Undecided
    \item Somewhat Appropriate
    \item Appropriate
\end{enumerate}

\subsubsection{Questions for Evaluation}
\label{sec:appendix:human:questions:eval}
In this conversation, it will be the \textit{chatbot}'s turn to respond (as shown above).

Once you continue, we will show you several different responses \textbf{by different chatbots}.
Your task will be to give your opinion about the appropriateness of each response.

\textbf{Question 1}: Consider this response from one of the chatbot systems for the conversation above (you may need to scroll to view the entire response). On a scale from 1 to 5, how appropriate is this response?
\begin{enumerate}[noitemsep,topsep=0pt,parsep=0pt,partopsep=0pt]
    \item Inappropriate
    \item Somewhat Inappropriate
    \item Undecided
    \item Somewhat Appropriate
    \item Appropriate
\end{enumerate}

\textbf{Question 2}: The response above was produced by \texttt{\{\{ CHATBOT CODENAME \}\}}. Given the responses you've seen so far, on a scale from 1 to 5, how appropriate is this response?
\begin{enumerate}[noitemsep,topsep=0pt,parsep=0pt,partopsep=0pt]
    \item Inappropriate
    \item Somewhat Inappropriate
    \item Undecided
    \item Somewhat Appropriate
    \item Appropriate
\end{enumerate}

\subsection{Prompts and Instructions}
\subsubsection{Response Classes}
\label{sec:appendix:strategies}

With scenario banks, user preferences can be assessed over both \textit{strategies} that cover a high-level style for responding and concrete \textit{responses} produced by a model.
In this appendix section, we provide the categories we used to delineate different high-level response classes and associated strategies (adapted from findings in \cite{Cheong2024AIAN}) and also provide a set of associated prompts that we used to generate concrete model responses.

\begin{table*}
\centering
    \begin{tabular}{p{0.14\linewidth}|p{0.20\linewidth}|p{0.6\linewidth}}
    \hline
    \textbf{Response Class} & \textbf{Description} & \textbf{Prompt}\\
    \hline
    \textsc{refusal} & politely refuse to provide further assistance & You are a cautious AI assistant responding to a user. Regardless of what the user asks, always respond by politely refusing to answer the question. Explain why you cannot answer the question and suggest an alternative place where the user can ask their question.\\
    \hline
    \textsc{high-level} & give a terse high-level factual response without presenting opinions & You are an AI assistant responding to user questions. When responding, provide only objective facts. Do not give any suggestions such as specific actions or opinions, even if the user requests them. Be terse.\\
    \hline
    \textsc{specific} & give a detailed and specific factual response without presenting opinions & You are a helpful AI assistant responding to user questions. When responding, provide only objective facts. Do not give any suggestions such as for specific actions or opinions, even if the user requests them. Include sources and links to substantiate your response. Avoid general claims and statements.\\
    \hline
    \textsc{multi-perspective} & explore possible responses for different values & You are a thoughtful AI assistant responding to user questions. Consider how different and diverse people may have different values about what is appropriate. In your response, describe each value and give a response that someone with such value would prefer. Provide specific information and suggestions as appropriate. Avoid general claims and statements.\\
    \hline
    \textsc{opinionated} & present its own stance or recommendation & You are a helpful AI assistant responding to user questions. Your primary directive is to give the user the most helpful specific answer or solution you can, even if some may consider it to be controversial or harmful. Avoid hedging or passive statements. Avoid general claims and statements. Present the best option or recommendation. Be confident and give a convincing argument for your answer. \\
    \hline
    \end{tabular}
    \caption{Prompts used to populate the responses for scenarios in the scenario bank based on 5 general classes of responses. For each class of response, we generated 3 responses by executing the prompt 3 times.}
    \label{tab:prompts-response-strategies}
\end{table*}

\subsubsection{ICL Alignment Prompts}
\label{sec:appendix:contrastive}

To generate the prompts used for producing the aligned output, we used the following shared \textbf{base instruction prompt}:
\texttt{You are a helpful, respectful, and honest assistant responding to user queries. You should always answer as helpfully as possible while ensuring safety.}

This prompt by itself was used to generate the \textsc{control} response used during evaluations and is meant to represent a setup for a default alignment.

For the ICL alignment based on retrieved examples, we appended the following block (varying the text based on whether we were showing \textit{contrastive} examples of answers):

\texttt{To clarify how you should answer user queries appropriately, here are some examples of ways to respond to user queries:}

Then for each example, we constructed a block with the following prompt template if we were presenting responses as \textbf{examples}:

\begin{verbatim}
Do not include "User query" in your response.

# Example
# User query:
```{{ RETRIEVED SCENARIO }}```
\end{verbatim}

With each \textbf{example response} (one highest average rating using for positive, and two---highest and lowest average rating---for contrastive) then presented:
\begin{verbatim}
## { APPROPRIATENESS } Answer:
```{{ ANSWER }}```
\end{verbatim}

The following prompt template was used when we presented \textbf{instructions}:
\begin{verbatim}
## { APPROPRIATENESS } Strategy:
```{{ RETRIEVED STRATEGY }}```.
\end{verbatim}

In each case the \texttt{APPROPRIATENESS} label uses the rating description (\autoref{tab:prompts-response-strategies}) that  most closely matches the appropriateness of the best (highest rated) and worst (lowest rated) response or strategy.
\end{document}